%% file: template.tex
\documentclass{article}

\usepackage{arxiv}

\usepackage[utf8]{inputenc} 
\usepackage[T1]{fontenc}    
\usepackage{hyperref}       
\usepackage{url}            
\usepackage{booktabs}       
\usepackage{amsfonts}       
\usepackage{nicefrac}       
\usepackage{microtype}      
\usepackage{lipsum}		
\usepackage{graphicx}
\usepackage{natbib}
\usepackage{doi}
\usepackage{float}
\usepackage{multirow}
\usepackage{graphicx}
\usepackage{tabularx}

\title{ChatGPT vs State-of-the-Art Models: A Benchmarking Study in Keyphrase Generation Task}

\author{
  Roberto Martínez-Cruz \\
  Machine Learning Engineer \\
  Moody's Analytics \\
  Prague, Czech Republic\\
  \texttt{roberto.martinezcruz@moodys.com} \\
  \And
  Alvaro J. López-López,  José Portela \\
  Machine Learning Researcher \\
  Institute for Research in Technology\\
  ICAI School of Engineering\\
  Comillas Pontifical University\\
  28015 Madrid, Spain\\
  \texttt{\{allopez, jportela\}@comillas.edu}
}

\renewcommand{\shorttitle}{\textit{arXiv} Template}

\hypersetup{
pdftitle={ChatGPT vs State-of-the-Art Models: A Benchmarking Study in Keyphrase Generation Task},
pdfsubject={cs.CL},
pdfauthor={Roberto Martínez Cruz},
pdfkeywords={ChatGPT, text generation, keyphrase generation, natural language processing, deep learning, domain adaptation, long documents},
}

\begin{document}
\maketitle

\begin{abstract}
	\input{abstract}
\end{abstract}


\section{Introduction}
\input{introduction}

\section{Related Work}
\input{related_work}

\section{Experimental Setup}
\label{sec:experimental_setup}
\input{experimental_setup}

\section{Results}
\input{results}

\section{Case Studies}
\input{case_studies}

\section{Conclusions and Future Work}
\input{conclusions}

\section{Acknowledgments}
We would like to express our gratitude to Debanjan Mahata, who served as our master and introduced us to the field of NLP and KPE. His guidance and patience have been invaluable throughout this research project, and we are grateful for his mentorship and support.

We would also like to thank Alejandro Pérez and Sergio Gago for providing the computational resources that were essential for developing and testing the ideas presented in this paper. His generosity and support have been instrumental in the success of this research project.

Finally, we would like to acknowledge the countless individuals and organizations who have contributed to the field of NLP and Keyphrase Extraction, as their work has provided the foundation for this research. We are grateful for their ongoing efforts and dedication to advancing this field, and we hope that this paper will contribute to their ongoing work.

Thank you all for your contributions and support.

\bibliographystyle{unsrtnat}
\bibliography{references}  






\end{document}

%% file: abstract.tex
Transformer-based language models, including ChatGPT, have demonstrated exceptional performance in various natural language generation tasks. However, there has been limited research evaluating ChatGPT's keyphrase generation ability, which involves identifying informative phrases that accurately reflect a document's content. This study seeks to address this gap by comparing ChatGPT's keyphrase generation performance with state-of-the-art models, while also testing its potential as a solution for two significant challenges in the field: domain adaptation and keyphrase generation from long documents. We conducted experiments on six publicly available datasets from scientific articles and news domains, analyzing performance on both short and long documents. Our results show that ChatGPT outperforms current state-of-the-art models in all tested datasets and environments, generating high-quality keyphrases that adapt well to diverse domains and document lengths.

%% file: introduction.tex
Keyphrase generation (KPG) is the process of automatically identifying or creating a set of phrases that effectively capture the essence of a document or text. These keyphrases provide a succinct summary of the main topics or themes discussed in the text and can be utilized for a range of downstream tasks such as document classification (\cite{kp-class}), clustering (\cite{kp-cluster}), summarization (\cite{kp-summarization}), recommendation (\cite{kp-recom}), and information retrieval (\cite{kp-information-retrieval}).

There are two types of keyphrases: extractive (found in the document) and abstractive (not found in the document). Historically, extractive methods, known as Keyphrase Extraction (KPE), based on sequence tagging models (\cite{nguyen_kan_2007,Gollapalli_Li_Yang_2017,alzaidy_2019, transkp,sahrawat2020keyphrase}) have demonstrated the highest accuracy, although they are limited in their ability to predict abstractive keyphrases. KPG offers two primary advantages over KPE: it can predict both extractive and abstractive keyphrases and It can leverage prompt-based learning to more effectively benefit from multitask learning.

The majority of KPG models follow the text-to-text generation training paradigm, which has greatly benefited from the development of transformer models and pre-trained language models (PLMs). These PLMs are capable of acquiring a comprehensive contextualized representation of the text by undergoing pre-training with a vast corpus of data via diverse self-supervised learning tasks. The current state-of-the-art (SotA) model, KeyBART (\cite{keybart}), is based on this training paradigm and employs the transformer-based architecture of BART (\cite{lewis2019bart}), a PLM, as its foundation. KeyBART has demonstrated promising performance on various keyphrase generation benchmarks, surpassing previous state-of-the-art models. However, there is still ample room for improvement in the KPG domain, particularly in the generation of abstractive keyphrases and the effective integration of external knowledge sources.

Text-to-text generation has proven to be an effective approach to facilitate multi-task learning. The incorporation of prompt-based learning, as demonstrated in models such as T5 (\cite{raffel2020exploring}), has significantly enhanced the performance of text-to-text generation models, particularly in few-shot and zero-shot learning scenarios.

Generative Pre-trained Transformer (GPT) models have garnered significant attention for their ability to generate coherent and context-aware text (\cite{Radford2018ImprovingLU}). The latest GPT model, ChatGPT (\cite{ouyang2022training}), based on GPT-3.5 (\cite{brown2020language}), has demonstrated its impressive ability to generate human-like responses to text-based conversations across various domains and topics. Given the adaptability of ChatGPT to new domains and the transfer learning capabilities from other tasks learned by the model, we believe that it has the potential to achieve state-of-the-art results in the KPG task. Additionally, ChatGPT may also help address a long-standing challenge in the field of KPG - domain adaptation. By leveraging the power of transfer learning and its ability to adapt to new domains, ChatGPT can potentially overcome domain-specific challenges in KPG and enhance the performance and efficiency of the task.

Keyphrase generation from long documents is a latent problem in the field. Current approaches often rely on summarizing texts, such as using abstracts, to identify important phrases. However, this method has limitations. Real-world situations may not always provide summaries, leading to reduced algorithm performance on longer texts. Additionally, crucial keyphrases may be missing from the summaries, and contextual information in the original text may not be reflected in them, greatly reducing the algorithm's effectiveness. While SotA approaches utilize contextualized text representations from PLMs, these representations are limited to a maximum number of words, preventing the embedding of long-term word relationships. The greater maximum token limit of ChatGPT, which is four times larger than that of KeyBART, may lead to better performance on long documents.

Our goal is to assess the potential of ChatGPT for KPG by conducting performance tests across diverse topics, such as scientific and news domains, as well as varying document lengths. We intend to determine whether ChatGPT's performance remains consistent across different document lengths and topics. To do so, we compare its results with the SotA algorithms, specifically KeyBART, on six publicly available datasets. Additionally, we provide real-world examples to demonstrate how ChatGPT generates keyphrases from input texts.

The contributions of this paper can be summarized as follows:
\begin{itemize}
    \item To the best of our knowledge, it’s the first attempt to employ ChatGPT for the KPG task in both the news and long document domains. To evaluate its performance, we compare it against state-of-the-art models across six widely used benchmark datasets. These datasets encompass both short and long scientific documents, as well as articles from the news domain.
    \item To the extend of our knoledge, no other study has benchmarked keyphrase generation from long documents. Previous studies have relied solely on the title and abstract to generate keyphrases, which may not provide an accurate representation of real-world scenarios where the entire document needs to be processed due to the absence of a summary.
    \item Our results are comprehensively analyzed, and case studies are presented to showcase the model's strengths and limitations in the KPG task. These case studies use real-world examples to demonstrate how the model leverages knowledge from other tasks and pretraining objectives to enhance keyphrase generation.
\end{itemize}

%% file: related_work.tex
\subsection{Keyphrase Extraction and Generation}
KPE involves selecting relevant phrases from a document, and there are two main approaches: supervised and unsupervised. Unsupervised methods typically use a two-step extraction process that involves heuristically identifying candidate phrases, then sorting and ranking them using graph-based approaches (\cite{textrank,topicrank,wang2014corpus,bennani2018simple,mahata2018key2vec}). Supervised methods use labeled data and can be customized for specific linguistic and contextual characteristics. Earlier supervised approaches relied on manually-engineered features (\cite{hulth2003,kim-kan-2009-examining,nguyen_kan_2007}), but a sequence labeling approach using a conditional random field (CRF) was introduced in \cite{Gollapalli_Li_Yang_2017}, and recent methods incorporate pre-trained word embeddings (\cite{alzaidy_2019}) like Word2Vec (\cite{word2vec}) or GloVe (\cite{glove}) to improve the accuracy. The improved sequence labeling approach uses a bidirectional long short-term memory (BiLSTM) + CRF layer to incorporate contextual information and model classification dependencies.

The transformer architecture, which has demonstrated improved performance in various natural language processing tasks, including as described in \cite{attention-is-all-you-need}, has been employed in several works for the KPE task, including TransKP (\cite{transkp}) and TNT-KID (\cite{tnt-kid}). Its ability to embed words in a sequence and provide representations that depend on the word and its context has led to the development of PLMs that specialize in providing contextualized embeddings. When combined with a trained BiLSTM-CRF layer, these embeddings have outperformed previous models, as demonstrated in \cite{sahrawat2020keyphrase}. Some works, such as KBIR (\cite{keybart}), have designed self-supervised objectives specifically for keyphrase extraction tasks to further enhance the representation of these embeddings, resulting in state-of-the-art results. 

However, the extractive model cannot handle the absent keyphrase. To overcome this limitation, KPG introduces a sequence-to-sequence generation method, first presented by \cite{meng-etal-2017-deep} through CopyRNN, a Seq2Seq framework that employs attention and copy mechanism. Since then, researchers have proposed several enhancements to this methodology. \cite{ye-wang-2018-semi} explored a semi-supervised method, \cite{chen2018keyphrase} investigated a review mechanism to reduce duplicates, and \cite{title-guided-kpg} focused on leveraging title information. Meanwhile, \cite{wang-etal-2019-topic-aware} exploited deeper topics of the document, and \cite{zhao-zhang-2019-incorporating} utilized linguistic constraints. Reinforcement learning was introduced by \cite{chan2019neural} and \cite{swaminathan-etal-2020-preliminary}. \cite{chen-etal-2020-exclusive} proposed an exclusive hierarchical decoding framework, while \cite{yuan-etal-2020-one} introduced a model that generates multiple keyphrases as delimiter-separated sequences. \cite{zhao2021sgg} proposed separate mechanisms to deal with present and absent keyphrase generation. \cite{Huang_Xu_Jiao_Zu_Zhang_2021} presented an AdaGM method to increase the discreteness of keyphrase generation, and \cite{ye-etal-2021-one2set} proposed an one2set method for generating diverse keyphrases as a set. Other works, including \cite{Chen2019AnIA}, \cite{ahmad-etal-2021-select}, and \cite{wu2021unikeyphrase}, focused on jointly learning extraction and generation for keyphrase prediction. \cite{Wu_Ma_Liu_Chen_Nie_2022} introduced prompt-based learning with a non-autoregressive approach, which is constrained to generate absent keyphrases.

The SotA results were achieved by KeyBART (\cite{keybart}), which presented a new pre-training technique for BART (\cite{lewis2019bart}). Unlike earlier pre-training methods that aimed to remove noise from the input text, KeyBART generates keyphrases related to the input text in concatenated sequential (CatSeq) format.

Keyphrase extraction and generation are crucial tasks in natural language processing, but only a few studies have addressed the challenge of extracting keyphrases from lengthy documents. One such study by \cite{mahata2022ldkp} released two large datasets containing fully extracted text and metadata, evaluating the performance of unsupervised and supervised algorithms for keyphrase extraction. Another notable example is the work by \cite{query-based-kpe}, which proposes a system that chunks documents while maintaining a global context as a query for relevant keyphrase extraction. They employ a pre-trained BERT model to estimate the probability of a given text span forming a keyphrase and find that a shorter context with a query outperforms a longer context without a query. \cite{martinezcruz2023} introduced a specialized approach to enhance keyphrase extraction from lengthy documents. They employed graph embedding techniques on the co-occurrence graph derived from the entire document, enriching the understanding of the document by incorporating a holistic perspective into the Pre-trained Language Model (PLM). The observed improvements underscore the importance of considering a comprehensive view of the full document for effective keyphrase extraction and generation. To enhance keyphrase generation, \cite{garg2022keyphrase} investigated the inclusion of information beyond the title and abstract as input in the field of keyphrase generation. Their approach demonstrated improved results, indicating that the model should not solely rely on the summary provided by the title and abstract to predict high-quality keyphrases. These studies highlight the significance of developing effective methods for keyphrase extraction and generation from lengthy documents and offer promising directions for future research.

While previous studies, such as \cite{song2023chatgpt}, have examined the effectiveness of ChatGPT in the KPG task, they did not assess its performance in full-length documents that exceed the model's maximum input limit without truncation, nor did they investigate its suitability for the news domain. Our study, on the other hand, provides a comprehensive analysis that extensively explores the capabilities of both KeyBART and ChatGPT across various use cases in the KPG task.

\subsection{GPT Models}
GPT models have been widely used in NLP tasks, such as language generation, language understanding, and question answering. GPT models, including GPT-2  (\cite{Radford2019LanguageMA}), GPT-3 and GPT 3.5 (\cite{brown2020language}), have achieved SotA performance in several NLP tasks and have become the de-facto standard in many applications. They exclusively use the transformer’s decoder and are pre-trained in a massive corpora of data using self-supervised learning.

Reinforcement learning from human feedback (RLHF) allows language models to learn from explicit feedback provided by human annotators, leading to improved text quality. Originally developed for training robots (\cite{NIPS2017_d5e2c0ad,ibarz2018reward}), recent studies have shown the benefits of applying RLHF to fine-tune language models (\cite{ziegler2020finetuning,stiennon2022learning,böhm2019better,wu2021recursively,jaques2019way,Kreutzer2018CanNM,lawrence-riezler-2018-improving,zhou2020learning,cho2019coherent,Perez2019FindingGE,madaan2023memoryassisted}), including GPT models such as ChatGPT (\cite{ouyang2022training}).

Multi-task learning has proven to be advantageous for GPT models since it entails instructing language models and is connected to cross-task generalization research in language models. Research has shown that fine-tuning language models on various NLP tasks with instructions can enhance their performance downstream, making it a powerful method for few-shot and zero-shot learning. These advantages have been corroborated in several studies (\cite{howard-ruder-2018-universal},\cite{devlin2019bert},\cite{dong2019unified},
\cite{mccann2018natural},\cite{keskar2019unifying}). 

By leveraging both RLHF and multi-task learning paradigms, ChatGPT has been fine-tuned from the GPT3.5 model to excel in chatbot development, surpassing its predecessors and showcasing its potential to revolutionize the field of conversational AI. While previous studies have benchmarked its performance in various NLP tasks, such as machine translation (\cite{hendy2023good}), there have been no previous studies exploring its potential in the KPG task.

%% file: experimental_setup.tex
\subsection{Datasets}
To evaluate the performance of KPG models, we employ the test set of six publicly available datasets covering both scientific and news domains, with varying document lengths. The datasets we use are as follows:

\begin{itemize}
    \item Inspec\footnote{\url{https://huggingface.co/datasets/midas/inspec}} (\cite{hulth2003}) is a scientific literature dataset that consists of 2,000 abstracts and their corresponding keyphrases, covering various topics The abstracts are from papers belonging to the scientific domains of Computers and Control and Information Technology published between 1998 to 2002. The dataset has a train,val and test split that contains 1000, 500 and 500 samples respectively.
    \item KP20K\footnote{\url{https://huggingface.co/datasets/midas/kp20k}} (\cite{meng-etal-2017-deep}), a large-scale dataset with over 528K articles for training, 20K articles for validation, and 20K articles for testing from the PubMed Central Open Access Subset, covering various domains including medicine, biology, and physics.
    \item The NUS\footnote{\url{https://huggingface.co/datasets/midas/nus}} dataset (\cite{nguyen_kan_2007}) consists of 211 full scientific documents that have been manually annotated with their respective keyphrases. It is used exclusively for evaluation purposes, as the dataset comprises solely a test split.
    \item SemEval2010\footnote{\url{https://huggingface.co/datasets/midas/semeval2010}} (\cite{kim_2010}), a dataset comprising 284 English full scientific papers from the ACM Digital Library, which are split into test and train sets containing 100 and 144 articles, respectively.
    \item The KPTimes\footnote{\url{https://huggingface.co/datasets/midas/kptimes}} (\cite{gallina2019kptimes}) dataset consists of 279,923 news articles from NY Times and 10K from JPTimes, curated by expert editors, and divided into train, validation, and test sets with 259,923, 10,000, and 20,000 samples, respectively.
    \item The DUC2001\footnote{\url{https://huggingface.co/datasets/midas/duc2001}} dataset (\cite{duc2001}) is a widely recognized corpus of news articles that includes 308 documents and 2,488 manually annotated keyphrases. It should be noted that this dataset only contains a test split and no training data.
\end{itemize}

Relevant statistics from the datasets can be found in Table \ref{tab:dataset_stats}.

\begin{table}[]
\centering
\begin{tabular}{@{}ccccccc@{}}
\toprule
\textbf{Dataset} & \textbf{Test Size} & \textbf{Long Doc} & \textbf{Domain} & \textbf{Avg. Words} & \textbf{Avg. Extractive KPs} & \textbf{Avg. Abstractive KPs} \\ \midrule
Inspec           & 500                & No                & Scientific      & 135                  & 6.56                          & 3.26                           \\ \midrule
KP20k            & 20,000             & No                & Scientific      & 160                  & 2.34                          & 2.94                           \\ \midrule
NUS              & 211                & Yes               & Scientific      & 9287                 & 8.02                          & 3.08                           \\ \midrule
SemEval2010      & 100                & Yes               & Scientific      & 8404                 & 9.17                          & 6.07                           \\ \midrule
KPTimes          & 20,000             & No                & News            & 643                  & 2.72                          & 2.3                            \\ \midrule
DUC2001          & 308             & No                & News            & 847                  & 7.14                          & 0.92                            \\ \bottomrule
\end{tabular}
\vspace*{2mm}
\caption{Statistics of the test splits for the dataset used in our experiments}
\label{tab:dataset_stats}
\end{table}

\subsection{Baselines}
The evaluation of ChatGPT's performance involved comparing it with several other generative models, with respect to their ability to predict missing keyphrases across different generation frameworks. These models are:

\begin{itemize}
    \item KeyBART\footnote{\url{https://huggingface.co/bloomberg/KeyBART}} (\cite{keybart}), which is a state-of-the-art model developed using BART and featuring a novel pre-training approach that generates relevant keyphrases in the CatSeq format, instead of just removing noise from input text.
    \item Prompt Based KPG (\cite{Wu_Ma_Liu_Chen_Nie_2022}), which utilizes a prompt-based learning method to generate missing keyphrases. The prompt is created based on the overlapping words between the absent keyphrase and the document, and a mask predict decoder is used to complete the keyphrase while adhering to the constraints of the prompt.
    \item UniKeyphrase (\cite{wu2021unikeyphrase}), which is a unified framework for both present keyphrase extraction and absent keyphrase generation. This framework is based on a pre-trained prefix LM model.
    \item Pure generative models that use a Seq2Seq approach including CatSeq (\cite{yuan-etal-2020-one}) and its enhanced versions such as CatSeqCorr (\cite{chen2018keyphrase}), catSeqTG (\cite{title-guided-kpg}), and CatSeqD (\cite{yuan-etal-2020-one}), along with ExHiRD-h model (\cite{chen-etal-2020-exclusive}), for comparison.
\end{itemize}

Relevant information from the main models can be found in Table \ref{tab:models_stats}.

\begin{table}[]
\centering
\begin{tabular}{@{}ccc@{}}
\toprule
\textbf{Model}  & \textbf{Training Domain} & \textbf{Max Input Tokens} \\ \midrule
ChatGPT         & MultiDomain              & 4,096  (Summing Input and Output Tokens)                    \\ \midrule
KeyBART         & Scientific              & 1024                     \\ \midrule
Prompt Base KPG & Scientific               & 384                       \\ \midrule
UniKeyphrase    & Scientific               & 384                       \\ \bottomrule
\end{tabular}
\vspace*{2mm}
\caption{Insights from the main models used in our experiments}
\label{tab:models_stats}
\end{table}

\subsection{Evaluation Metrics}

We used the $F1@K$ evaluation metric (\cite{kim_2010}), where $K$ represents the number of predicted keyphrases to be considered. Equations \ref{precision}, \ref{recall}, and \ref{f1} illustrate how to compute $F1@K$. Prior to evaluation, we preprocessed the ground truth and predicted keyphrases by converting them to lowercase, stemming them, and removing punctuation, and we used exact matching. Let $Y$ denote the ground truth keyphrases, and $\bar{Y} = (\bar{y_1},\bar{y_2}, \dots, \bar{y_m})$ denote the predicted keyphrases. The metrics are defined as follows:

\begin{equation}
    Precision@k = \frac{|Y \cap \bar{Y_k} |}{min\{\bar{|Y_k|},k\}}
\label{precision}
\end{equation}
\vspace*{1mm}

\begin{equation}
    Recall@k = \frac{|Y \cap \bar{Y_k} |}{|Y|}
\label{recall}
\end{equation}
\vspace*{1mm}

\begin{equation}
    F1@k = \frac{2*Precision@k*Recall@k}{Precision@k + Recall@k}
\label{f1}
\end{equation}
\vspace*{2mm}

Here, $\bar{Y}_k$ denotes the top $k$ elements of the predicted keyphrase set $\bar{Y}$. In our case, we set $K$ to the total number of predicted keyphrases $M$, and $5$ to represent the top 5 keyphrases generated by the model.

\subsection{Setting}
For KeyBART, the generated results are produced using a beam search with a beam width of 50, and the maximum length of the generated sequence is restricted to 40 tokens with a temperature of 0. When the number of input tokens exceeds the model's maximum input limit, the tokens are divided into non-overlapping chunks of the same size as the maximum input token limit, and the generated keyphrases are produced accordingly, concatenated and any duplicates are removed.

\begin{figure}[]
    \centering
    \includegraphics[width=1\textwidth]{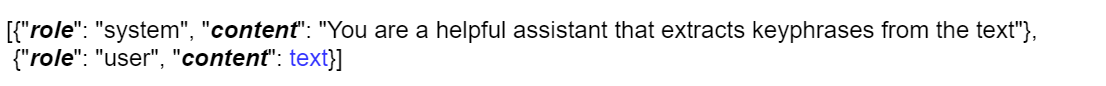}
    \caption{Example prompt used to generate keyphrases with ChatGPT}
    \label{fig:system_user_interaction}
\end{figure}

To generate keyphrases using ChatGPT, we utilize the gpt-3.5-turbo model in chat completion mode, with the prompt illustrated in Figure \ref{fig:system_user_interaction}. In the prompt, the $text$ variable represents the input text from which the keyphrases are to be generated. The generated sequence is restricted to 40 tokens with a temperature of 0, a frequency penalty of 0.8 and a presence penalty of 0. Although the model's response may vary in the chat format, it always includes a list of keyphrases presented in one of three ways: comma-separated, enumerated, or itemized. The response can be transformed into a list of keyphrases by post-processing it using regular expressions. Since the experiments involve processing long documents that exceed the maximum input token limit, the text is split into non-overlapping 2000-word segments and input using individual prompts. The final step involves concatenating the results and removing any duplicates to generate the complete list of keyphrases. 

The primary aim of this study is to evaluate the performance of scientific domain models using datasets consisting of short documents. Unlike prior studies that relied on the title and abstract of papers in long document datasets, we will use the entire documents to assess the models' effectiveness on longer texts. Moreover, we will compare the models' ability to adapt to different domains by evaluating their performance on news domain datasets. To conduct these two analyses, we will compare the results of the current SotA model, KeyBART, and ChatGPT.

We used the first five generated keyphrases for $F1@5$ to evaluate the models. If there were fewer than five keyphrases, we appended random incorrect keyphrases until there were five predictions. However, due to multiple separate generations for the same sample in long documents, this metric did not provide clear insights into the model's performance. Therefore, in this scenario we only benchmarked the results based on $F1@M$. In order to compare the ground truth keyphrases with the generated ones, we utilize the Porter Stemmer to normalize both as \cite{meng-etal-2017-deep}.

%% file: results.tex
This section presents the results of our experiments. ChatGPT outperforms the specialized models in the task in all scenarios, with the performance improvement becoming more significant as the document length increases or as the document's domain moves farther away from the scientific domain.

\subsection{Short Scientific Documents}
As previously highlighted, Inspec and KP20k datasets are used to benchmark the performance of the models in the short scientific documents domains. Scenario in which all the models are specialized with the exception of Chat GPT. In Table \ref{tab:short-docs-results} the results are shown.

\begin{table}[]
\centering
\resizebox{0.8\textwidth}{!}{%
\begin{tabular}{|c|l|cc|cc|}
\hline
\multirow{2}{*}{Task}                & \multicolumn{1}{c|}{\multirow{2}{*}{Model}} & \multicolumn{2}{c|}{Inspec}     & \multicolumn{2}{c|}{KP20k}      \\
                                     & \multicolumn{1}{c|}{}                       & F1@5           & F1@M           & F1@5           & F1@M           \\ \hline
\multirow{10}{*}{Present Keyphrases} & catSeq                                      & 0,225          & 0,262          & 0,291          & 0,367          \\
                                     & CatSeqD                                     & 0,219          & 0,263          & 0,285          & 0,363          \\
                                     & CatSeqCorr                                  & 0,227          & 0,269          & 0,289          & 0,365          \\
                                     & CatSeqTG                                    & 0,229          & 0,270          & 0,292          & 0,366          \\
                                     & ExHiRD-h                                    & 0,253          & 0,291          & 0,311          & 0,364          \\
                                     & SEG-Net                                     & 0,216          & 0,265          & 0,311          & 0,379          \\
                                     & UniKeyphrase                                & 0,260          & 0,288          & 0,347          & 0,352          \\
                                     & Prompt Base KPG                             & 0,260          & 0,294          & \textbf{0,351} & 0,355          \\
                                     & KeyBART                                     & 0,278          & 0,301          & 0,301          & \textbf{0,398} \\
                                     & ChatGPT                                     & \textbf{0,352} & \textbf{0,403} & 0,232          & 0,251          \\ \hline
\multirow{10}{*}{Absent Keyphrases}  & catSeq                                      & 0,004          & 0,008          & 0,015          & 0,032          \\
                                     & CatSeqD                                     & 0,006          & 0,011          & 0,015          & 0,031          \\
                                     & CatSeqCorr                                  & 0,005          & 0,009          & 0,015          & 0,032          \\
                                     & CatSeqTG                                    & 0,005          & 0,011          & 0,015          & 0,032          \\
                                     & ExHiRD-h                                    & 0,011          & 0,022          & 0,016          & 0,032          \\
                                     & SEG-Net                                     & 0,009          & 0,015          & 0,018          & 0,036          \\
                                     & UniKeyphrase                                & 0,012          & 0,022          & 0,032          & 0,058          \\
                                     & Prompt Base KPG                             & 0,017          & 0,022          & 0,032          & 0,042          \\
                                     & KeyBART                                     & 0,041          & 0,045          & 0,035          & 0,035          \\
                                     & ChatGPT                                     & \textbf{0,049} & \textbf{0,059} & \textbf{0,044} & \textbf{0,056} \\ \hline
\end{tabular}%
}
\vspace{3mm}
\caption{Results for short scientific documents of keyphrase prediction on benchmarks. The bold-faced values indicate the best performances across the board.}
\label{tab:short-docs-results}
\end{table}

Despite all models being specialized in the domain, ChatGPT surpasses them in all cases except for the KPE task in KP20K, where the fine-tuning of the models on this dataset explains their exceptional performance. However, ChatGPT outperforms them in the generation of abstractive keyphrases in both datasest. We speculate that this task benefits greatly from ChatGPT's multi-task learning, which enables the model to learn useful knowledge from other tasks and apply it to generate high-quality abstractive keyphrases; which could explain why the specialized models fine-tuned in the distribution were outperformed by ChatGPT.

\subsection{Long Scientific Documents}
To asses the performance of the models in long scientific documents the SemEval2010 and NUS datasets are utilized, where the number of tokens of each sample is several times larger than the maximum allowed by the models. The results of these experiments are presented in Table \ref{tab:long-docs-results}.

In this scenario, it is evident that ChatGPT outperforms KeyBART by a substantial margin. This can be credited to ChatGPT's higher input token limit, which enables it to capture more contextual information, such as distant word relationships. This is crucial for generating accurate and relevant keyphrases, giving ChatGPT a clear advantage over KeyBART in this task. Furthermore, we hypothesize that the incorporation of other tasks during training has facilitated the development of a larger language model without compromising its performance in this specific task. This approach could potentially be used to train larger language models specialized in KPG.

\begin{table}[]
\centering
\resizebox{0.6\textwidth}{!}{%
\begin{tabular}{|c|c|c|c|}
\hline
\multirow{2}{*}{Task}               & \multirow{2}{*}{Model} & SemEval2010    & NUS            \\
                                    &                        & F1@M           & F1@M           \\ \hline
\multirow{2}{*}{Present Keyphrases} & KeyBART                & 0,137          & 0,143          \\
                                    & ChatGPT                & \textbf{0,186} & \textbf{0,1996} \\ \hline
\multirow{2}{*}{Absent Keyphrases}  & KeyBART                & 0,019          & 0,010          \\
                                    & ChatGPT                & \textbf{0,021} & \textbf{0,042} \\ \hline
\end{tabular}%
}
\vspace{3mm}
\caption{Results for long scientific documents of keyphrase prediction on benchmarks. The bold-faced values indicate
the best performances across the board.}
\label{tab:long-docs-results}
\end{table}

\subsection{News Domain}
To evaluate the domain adaptation abilities of the models, we will employ the news domain datasets DUC2001 and KPTimes. These datasets exhibit significant differences in both their domain and distributions when compared to the scientific domain. However, both models have been pre-trained on this domain. KeyBART, for example, is built on BART, which includes this distribution in its pre-training. The results are displayed in Table \ref{tab:news-results}.

\begin{table}[]
\centering
\resizebox{0.7\textwidth}{!}{%
\begin{tabular}{|c|c|cc|cc|}
\hline
\multirow{2}{*}{Task}               & \multirow{2}{*}{Model} & \multicolumn{2}{c|}{DUC2001}    & \multicolumn{2}{c|}{KPTimes}    \\
                                    &                        & F1@5           & F1@M           & F1@5           & F1@M           \\ \hline
\multirow{2}{*}{Present Keyphrases} & KeyBART                & 0,023          & 0,079          & 0,023          & 0,063          \\
                                    & ChatGPT                & \textbf{0,267} & \textbf{0,292} & \textbf{0,279} & \textbf{0,290} \\ \hline
\multirow{2}{*}{Absent Keyphrases}  & KeyBART                & 0,001          & 0,001          & 0,006          & 0,010          \\
                                    & ChatGPT                & \textbf{0,029} & \textbf{0,030} & \textbf{0,021} & \textbf{0,022} \\ \hline
\end{tabular}%
}
\vspace{3mm}
\caption{Results for news of keyphrase prediction on benchmarks. The bold-faced values indicate
the best performances across the board.}
\label{tab:news-results}
\end{table}

As demonstrated by the results, ChatGPT achieved significantly stronger performance in this domain, surpassing the results of KeyBART by a factor of three or more in every benchmark. The notable performance difference between ChatGPT and KeyBART can be attributed to the fact that although KeyBART's weights include knowledge from the news domain due to BART's pre-training, this knowledge may have been relatively forgotten during KeyBART's pre-training and fine-tuning in the scientific domain. It's worth noting that ChatGPT may not have been explicitly trained on the news domain for the KPG task. However, during its training, other tasks in the news domain were included, and ChatGPT may have reused useful knowledge from these tasks to improve its performance on KPG.

%% file: case_studies.tex
This section presents several case studies, including both short and long scientific documents, as well as a news example. Our goal is to demonstrate how ChatGPT can enhance the Keyphrase Generation (KPG) task and highlight its benefits in real-world scenarios. For each sample, we used both the KeyBART and ChatGPT models to generate the keyphrases. To make it easier to view our results, we have color-coded the keyphrases in the accompanying images according to the scheme specified in Table \ref{tab:highlight-colors}. It is important to note that many of the keyphrases that were labeled as abstractive are actually extractive. Additionally, in short documents, most of the keyphrases labeled as abstractive appear later in the full document, underscoring the importance of processing the entire long document rather than just its summarized abstract.

\begin{table}[]
\centering
\begin{tabular}{@{}cc@{}}
\toprule
\textbf{Highlighting Color} & \textbf{Associated Significance}                   \\ \midrule
Yellow                      & True keyphrase unpredicted by both models          \\ \midrule
Light Blue                  & True keyphrase correctly predicted by both models  \\ \midrule
Dark Blue                   & True keyphrase correctly predicted by KeyBART only \\ \midrule
Green                       & True keyphrase correctly predicted by ChatGPT only \\ \bottomrule
\end{tabular}
\vspace{3mm}
\caption{Corresponding Significance of Each Highlight Color}
\label{tab:highlight-colors}
\end{table}

\subsection{Case Study 1: Short Scientific Documents}

\subsubsection{Case Study 1.1}
\begin{figure}[H]

    \centering
    \includegraphics[width=0.8\textwidth]{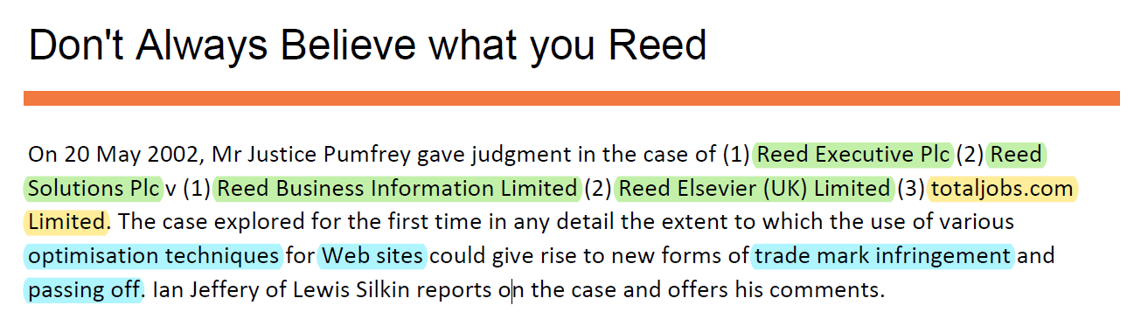}
    \caption{Case Study 1.1: Short Scientific Document from Inspec Test Dataset}
    \label{fig:case-study-1}
\end{figure}

This case study presents a short scientific document, specifically a sample of the Inspec test dataset with ID 2166, as shown in Figure \ref{fig:case-study-1}.

\begin{figure}[]
    \centering
    \includegraphics[width=0.7\textwidth]{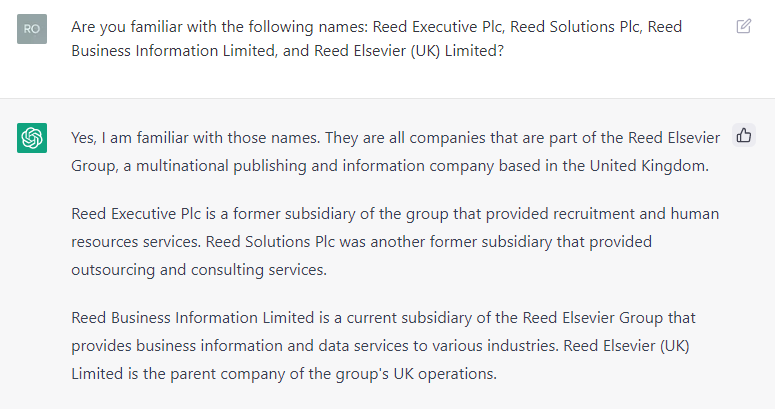}
    \caption{Case Study 1.1: ChatGPT's knowledge in the field directly questioned}
    \label{fig:case-study-1-chatgpt}
\end{figure}

As observed, KeyBART was unable to predict half of the KPs in its specialized domain. We hypothesize that due to the document's brevity, the model struggles to identify the keywords' significance. However, ChatGPT was not hindered by this limitation and utilized its pretraining knowledge to assign importance to these phrases. We queried ChatGPT directly using its front-end to determine if it was familiar with the keyphrases exclusively predicted by it, and the response can be seen in Figure \ref{fig:case-study-1-chatgpt}. The model not only recognized the terms but also grouped them together and provided a brief introduction, demonstrating how the model's knowledge from other tasks and pretraining objectives can be useful for KPG.

\subsubsection{Case Study 1.2}

\begin{figure}[]
    \centering
    \includegraphics[width=0.8\textwidth]{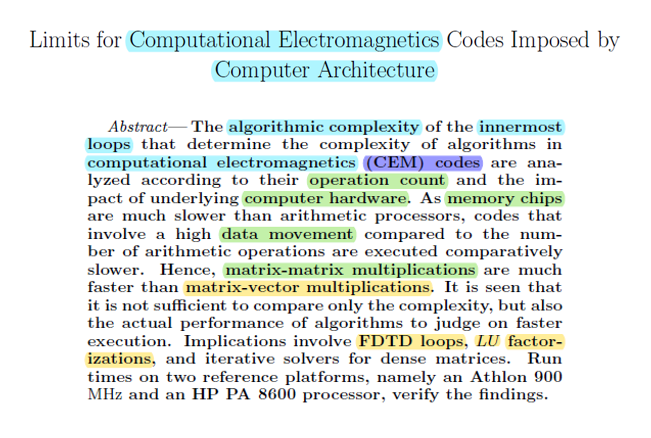}
    \caption{Case Study 1.2: Short Scientific Document from Inspec Test Dataset}
    \label{fig:case-study-1.2}
\end{figure}

The sample document chosen for this case study is a short scientific piece shown in Figure \ref{fig:case-study-1.2}, sourced from the Inspec test dataset and bearing the identification number 2043. 

Interestingly, ChatGPT exhibits superior performance in this abstract, which contains no repeated keyphrases. Such documents can be more challenging, as the importance of a word cannot be inferred from its redundant references. It is worth mentioning that these keyphrases are reiterated throughout the entire document, emphasizing the need for a comprehensive understanding of the full text to accurately identify keyphrases.

KeyBART's performance excels in the initial sections of the document but declines in the middle, and neither model successfully predicts any keyphrases in the latter parts. This observation suggests that the final sections are more complex, as the significance of a word cannot be deduced from its earlier mentions. This limitation impacts KeyBART more severely than ChatGPT.

\begin{figure}[]
    \centering
    \includegraphics[width=0.75\textwidth]{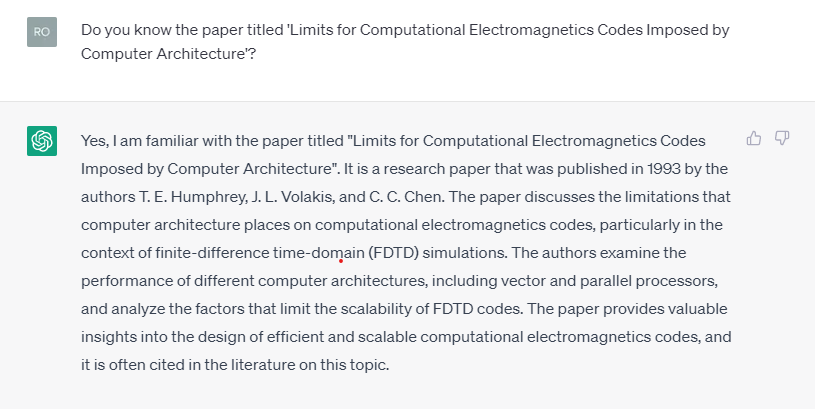}
    \caption{Case Study 1.2: Direct Inquiry of ChatGPT's Familiarity with the Paper}
    \label{fig:case-study-1.2-chatgt}
\end{figure}

In this instance, as demonstrated in Figure \ref{fig:case-study-1.2-chatgt}, we directly asked ChatGPT if it was familiar with the paper. Although the model could not correctly predict specific details from the document, such as the authors' names, it successfully summarized the paper's main points. This implies that ChatGPT possesses accurate latent knowledge of the article acquired during its pre-training, which in turn enhances its performance in the keyphrase generation task.

\subsubsection{Case Study 1.3}
\begin{figure}[]
    \centering
    \includegraphics[width=0.5\textwidth]{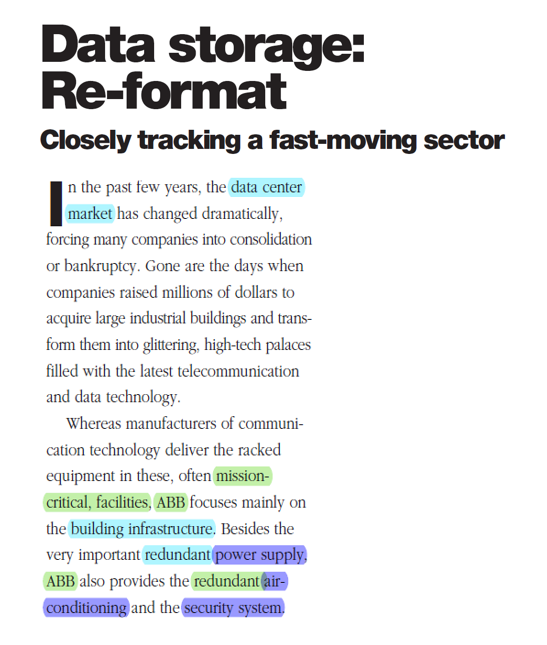}
    \caption{Case Study 1.3: Short Scientific Document from Inspec Test Dataset}
    \label{fig:case-study-1.3}
\end{figure}

This case study features another short scientific article, identified by number 2150 and sourced from the Inspec test dataset, as illustrated in Figure \ref{fig:case-study-1.3}.

KeyBART and ChatGPT display comparable performance in this case, with a low intersection between their correct predictions. KeyBART succeeds in accurately predicting generic phrases from the scientific domain, such as "power supply," whereas ChatGPT is more adept at identifying specific terms, like the company name "ABB," likely due to its industry knowledge. Interestingly, both models have several inaccurately predicted keyphrases in common, such as "Data Storage," which could potentially be a keyphrase that the author overlooked. Keyphrase identification can sometimes be ambiguous.

\begin{figure}[]
    \centering
    \includegraphics[width=0.75\textwidth]{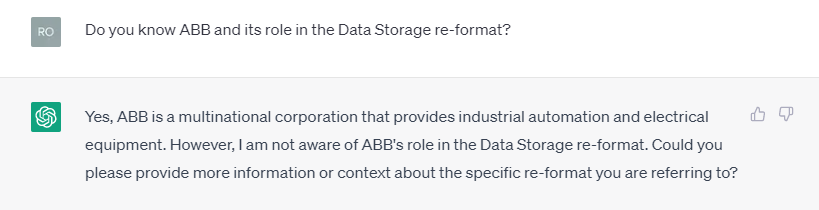}
    \caption{Case Study 1.3: Investigating ChatGPT's Familiarity with the Company and Topic Discussed in the Paper}
    \label{fig:case-study-1.3-chatgt}
\end{figure}

We asked the model directly about ABB, a company involved in data storage re-formatting. Figure \ref{fig:case-study-1.3-chatgt} displays the results, which show that while the model accurately describes the company, it lacks knowledge about the specific topic described in the article. This instance represents ChatGPT's lowest performance on the Inspec dataset. It seems that the model relies heavily on its inner knowledge of the document, which may explain its lower accuracy in cases where it lacks specific information.

\subsection{Case Study 2: Long Scientific Documents}
\subsubsection{Case Study 2.1}
\begin{figure}[]
    \centering
    \includegraphics[width=0.65\textwidth]{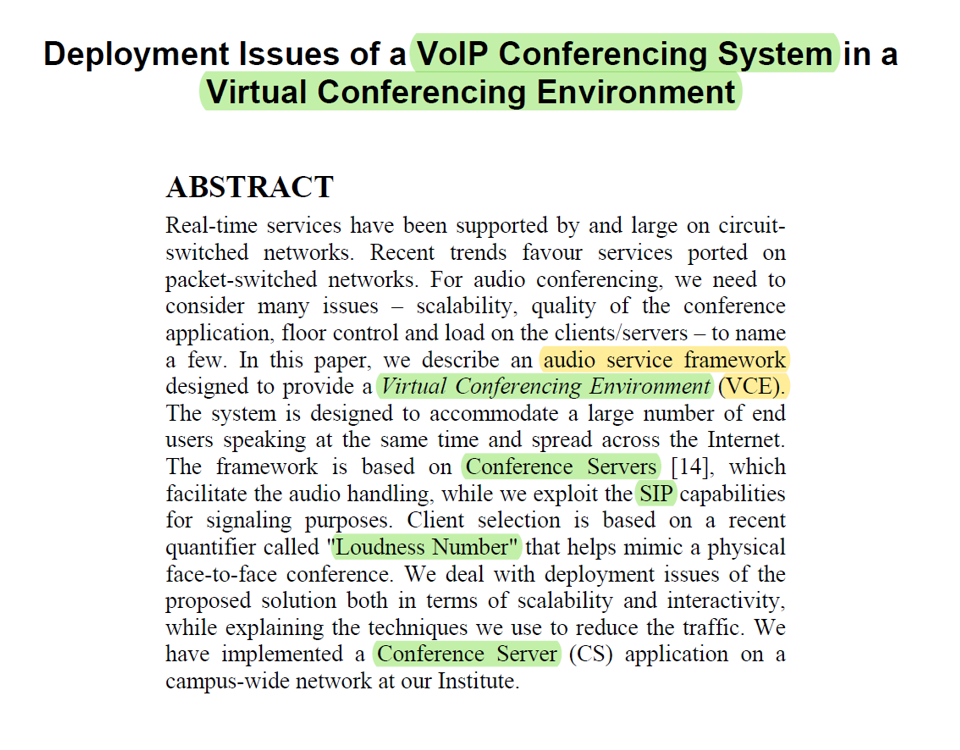}
    \caption{Case Study 2.1: Long Scientific Document from SemEval2010 Test Dataset - Title \& Abstract}
    \label{fig:case-study-2-abstract}
\end{figure}

This case evaluates the models' performance on long scientific documents, specifically a sample of the SemEval2010 test dataset with ID 'C-17'. Its title and abstract are shown in Figure \ref{fig:case-study-2-abstract}.

\begin{figure}[]
    \centering
    \includegraphics[width=0.75\textwidth]{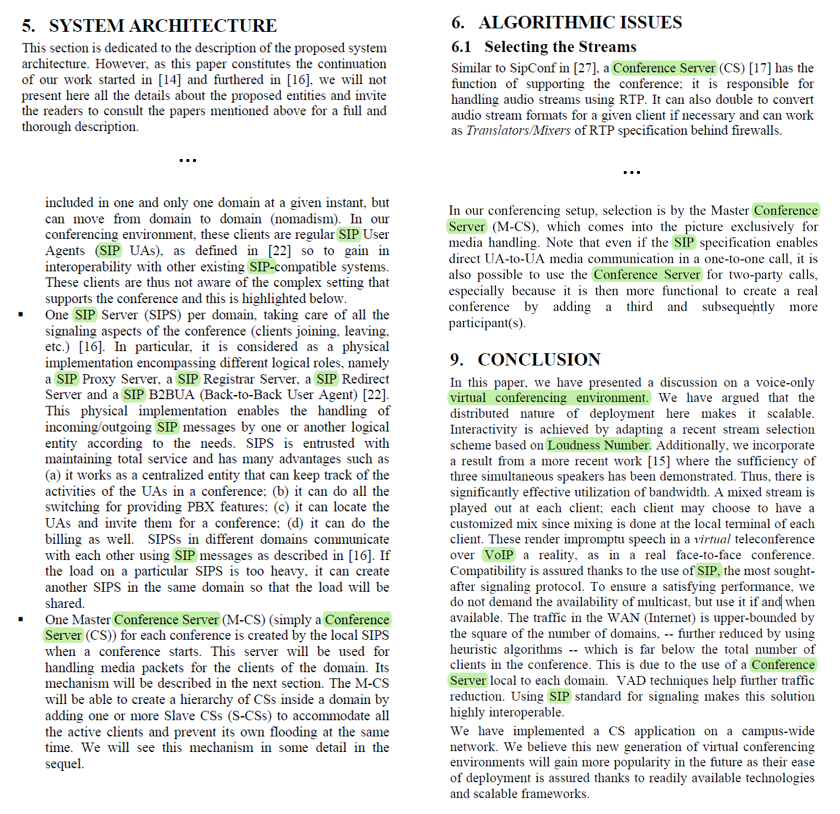}
    \caption{Case Study 2.1: Long Scientific Document from SemEval2010 Test Dataset - Parts of the Body \& Conclusion}
    \label{fig:case-study-2-conclusion}
\end{figure}

In this example, the lack of context in the title and abstract made it impossible for KeyBART to identify any keyword. However, due to its pre-training and field knowledge, it was still able to generate coherent abstractive keyphrases such as 'teleconferencing' and 'collaborative virtual environment.' In the case of long documents, the limitations posed by the maximum input tokens may prevent the model from comprehending long-term relationships between words necessary for generating keyphrases. The higher maximum input tokens in ChatGPT have contributed to its superior performance. As shown in Figure \ref{fig:case-study-2-conclusion}, keyphrases such as 'SIP' and 'Conference Server' gain greater importance in later parts of the document, such as the conclusion, which cannot be extrapolated without a holistic view of the entire document.

\begin{figure}[]
    \centering
    \includegraphics[width=0.8\textwidth]{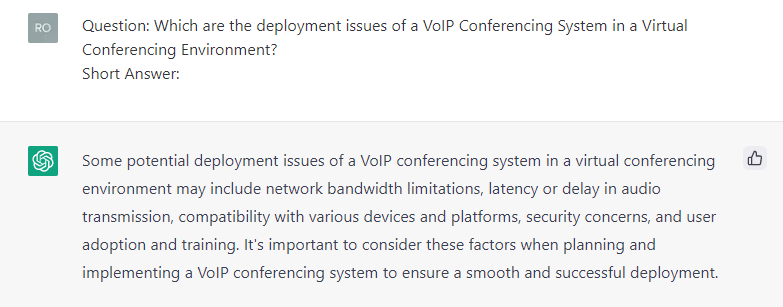}
    \caption{Case Study 2.1: ChatGPT's knowledge in the domain of the article}
    \label{fig:case-study-2-chatgpt}
\end{figure}

Domain knowledge gained from training on other tasks is another key component that explains ChatGPT's superior performance in such documents. Figure \ref{fig:case-study-2-chatgpt} illustrates how the model is capable of providing a coherent response when asked a question derived from the article's title, demonstrating the crucial role of domain knowledge in KPG from long articles.

\subsubsection{Case Study 2.2}
\begin{figure}[]
    \centering
    \includegraphics[width=0.65\textwidth]{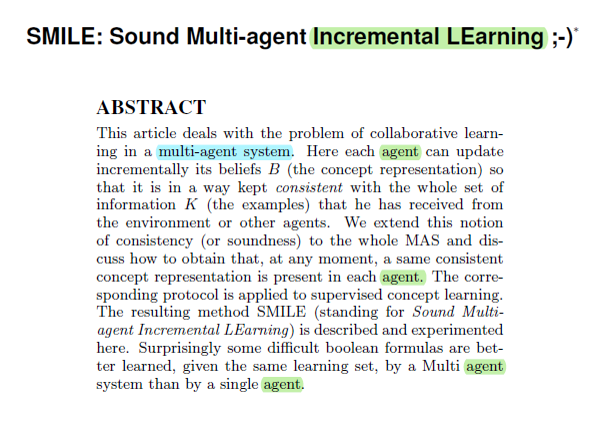}
    \caption{Case Study 2.2: Long Scientific Document from SemEval2010 Test Dataset - Title \& Abstract}
    \label{fig:case-study-2.2-abstract}
\end{figure}

This example pertains to an extensive scientific paper from the SemEval2010 test dataset, identified as 'I-10'. The title and abstract can be viewed in Figure \ref{fig:case-study-2.2-abstract}.

\begin{figure}[]
    \centering
    \includegraphics[width=1\textwidth]{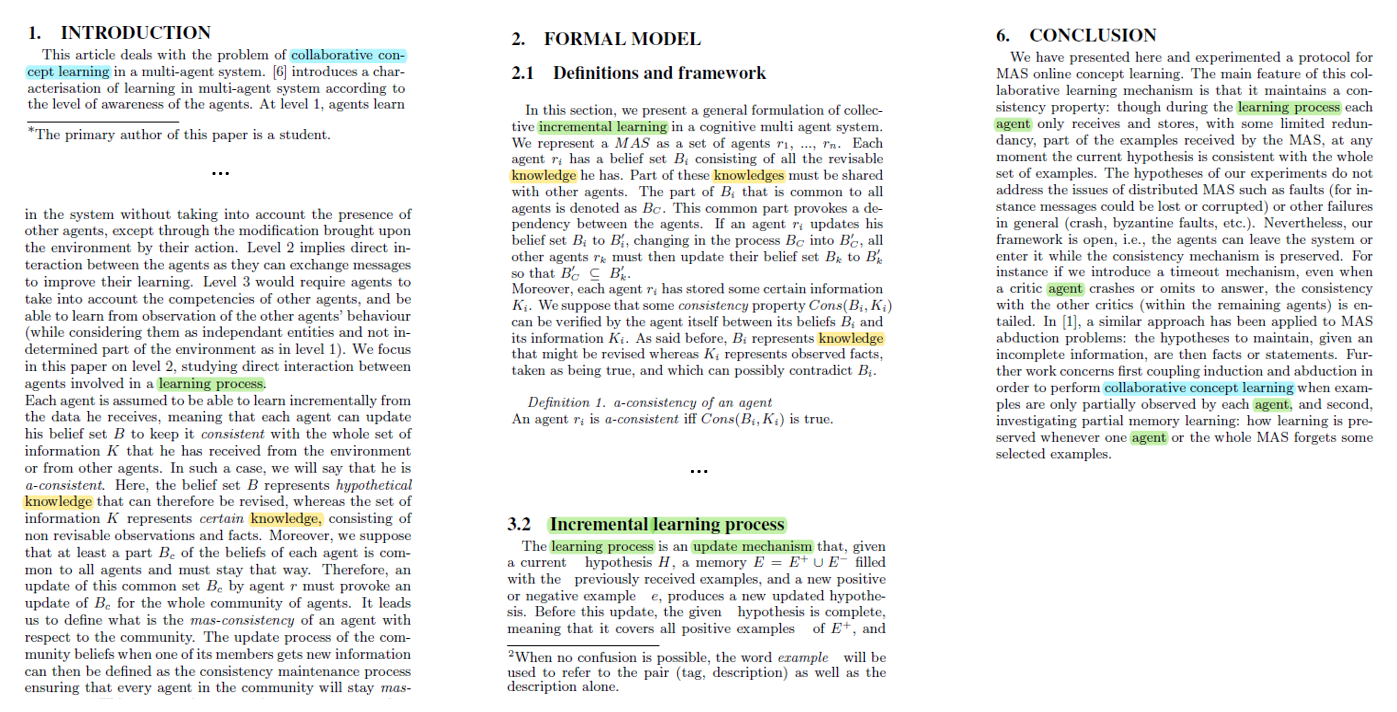}
    \caption{Case Study 2.2: Long Scientific Document from SemEval2010 Test Dataset - Parts of the Body \& Conclusion}
    \label{fig:case-study-2.2-intro}
\end{figure}

Observably, KeyBART struggles to identify 'incremental learning' from the title and fails to recognize the repeated term 'agent.' In contrast, ChatGPT accurately predicts both elements. Nevertheless, KeyBART does partially capture 'agent' with predictions such as 'multi-agent system' and 'autonomous agent.' Furthermore, it accurately anticipates the abstractive keyphrase 'mas-consistency' through alternative expressions like 'mas-consistent learning,' which ChatGPT did not manage to recognize.

Key portions of the document, as well as the conclusion, can be found in Figure \ref{fig:case-study-2.2-intro} Notably, both models failed to identify 'knowledge' as a keyphrase, despite its repetition more than 10 times throughout the document. This oversight could result from the models lacking full document context, causing them to perceive 'knowledge' as a common word. Another instance is KeyBART's inability to recognize 'Incremental Learning,' which appears in significant sections later in the document, such as a subsection title. This limitation might be related to its narrow token context or a lack of familiarity with the paper's specialized area.

\begin{figure}[]
    \centering
    \includegraphics[width=0.8\textwidth]{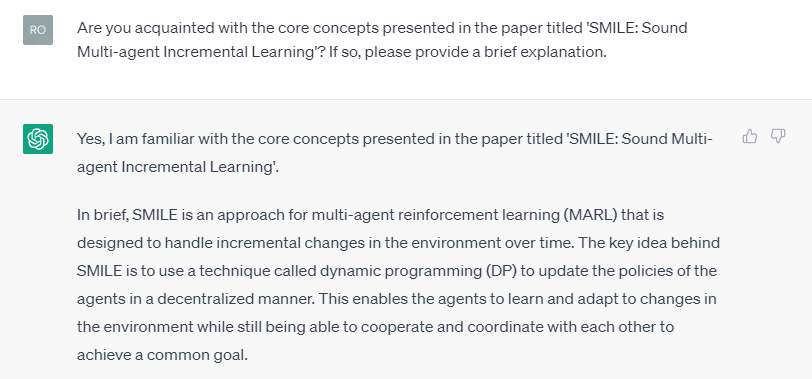}
    \caption{Case Study 2.2: ChatGPT's Familiarity with the Paper}
    \label{fig:case-study-2.2-chatgpt}
\end{figure}

\begin{figure}[]
    \centering
    \includegraphics[width=0.8\textwidth]{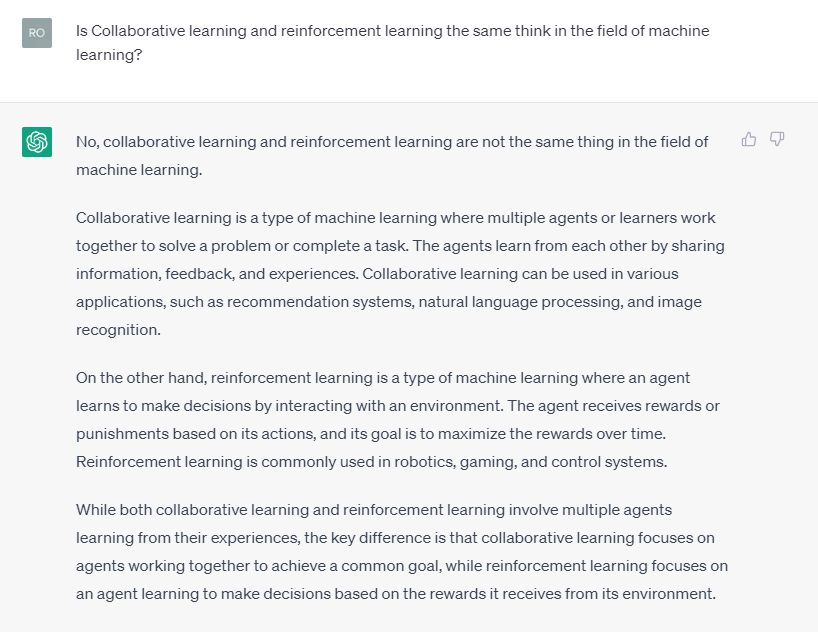}
    \caption{Case Study 2.2: Examining ChatGPT's Comprehension of Reinforcement and Collaborative Learning}
    \label{fig:case-study-2.2-chatgpt-2}
\end{figure}

In order to comprehend the implicit knowledge contained in the document produced by ChatGPT, we requested that it provide us with a brief summary of the key concepts it covers. The resulting summary is displayed in Figure \ref{fig:case-study-2.2-chatgpt}. However, the model incorrectly characterized the collaborative learning method outlined in the document as reinforcement learning. Despite this error, the model did correctly identify that the primary focus of the paper was artificial intelligence, indicating that its prediction was not entirely incorrect. Additionally, we directly queried the model regarding its understanding of the difference between collaborative learning and reinforcement learning, and the results are shown in \ref{fig:case-study-2.2-chatgpt-2}. The model demonstrated a clear understanding of the distinction between the two terms, suggesting that its misidentification of the paper's content may have been an informed guess. 

\subsubsection{Case Study 2.3}
\begin{figure}[]
    \centering
    \includegraphics[width=0.65\textwidth]{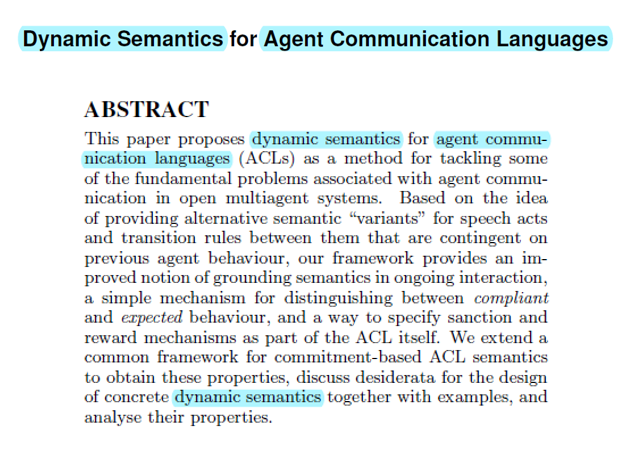}
    \caption{Case Study 2.3: Long Scientific Document from SemEval2010 Test Dataset - Title \& Abstract}
    \label{fig:case-study-2.3-abstract}
\end{figure}

This instance relates to a long scientific document in the SemEval2010 evaluation dataset, designated as 'I-10'. The title and abstract are observable in Figure \ref{fig:case-study-2.3-abstract}.

\begin{figure}[]
    \centering
    \includegraphics[width=1\textwidth]{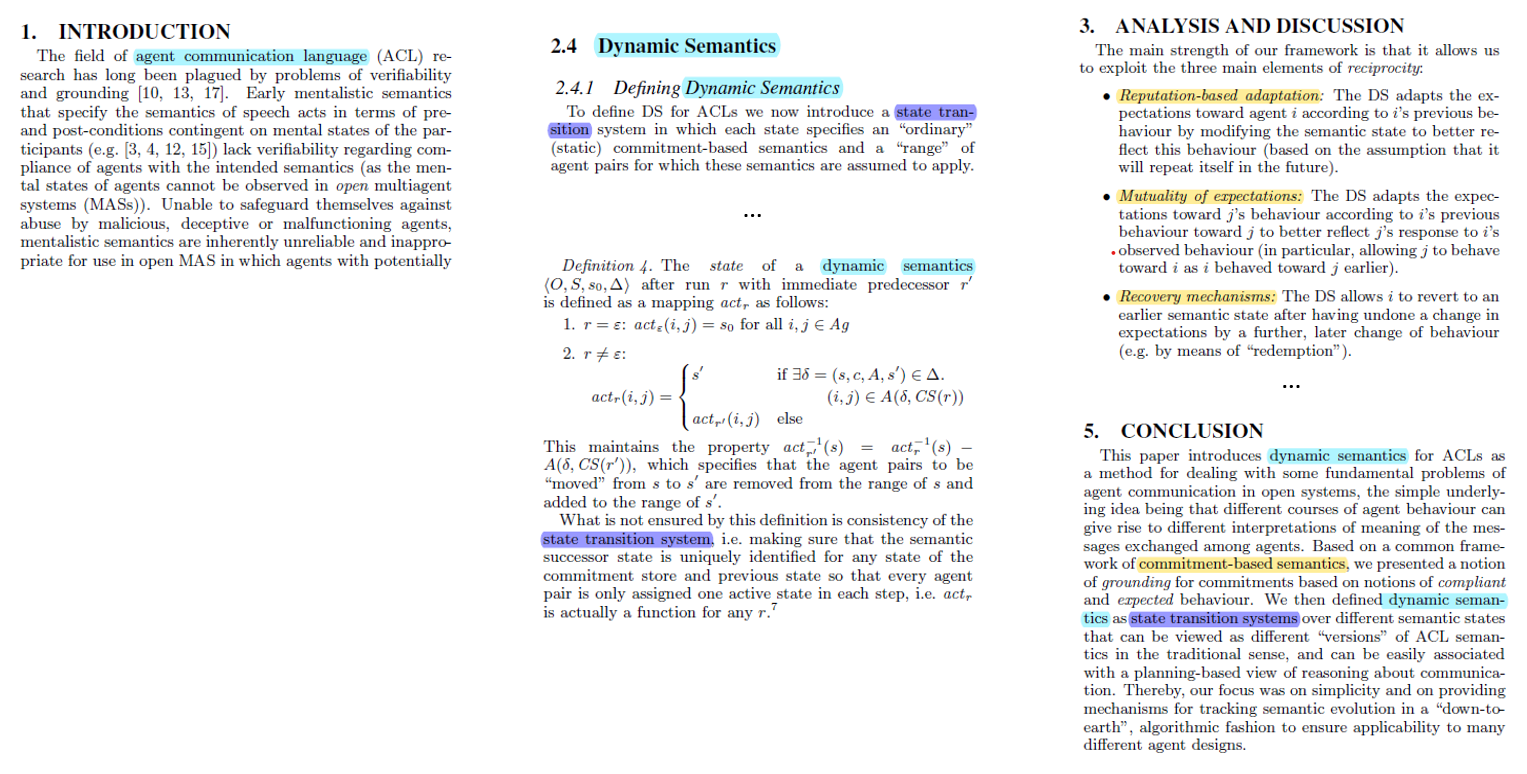}
    \caption{Case Study 2.3: Long Scientific Document from SemEval2010 Test Dataset - Parts of the Body \& Conclusion}
    \label{fig:case-study-2.3-intro}
\end{figure}

In this example, both models accurately identify the keyphrases covered in the document's title and abstract. However, as shown in Figure \ref{fig:case-study-2.3-intro}, neither can identify phrases located at crucial points in the document, such as 'Reputation-based adaption' or 'commitment-based semantics', the latter even being included in the conclusion. We speculate that this may be due to the lack of long-term relationship embedding mechanisms in both models.

It's worth noting that KeyBART outperforms ChatGPT in this sample. KeyBART can correctly identify all the keyphrases that ChatGPT predicts correctly and, in addition, correctly predicts the extractive keyphrase 'state transition systems' and comes close to correctly predicting the abstractive keyphrase 'social notion', with predictions such as 'social science' or 'social attitude'.

\begin{figure}[]
    \centering
    \includegraphics[width=0.8\textwidth]{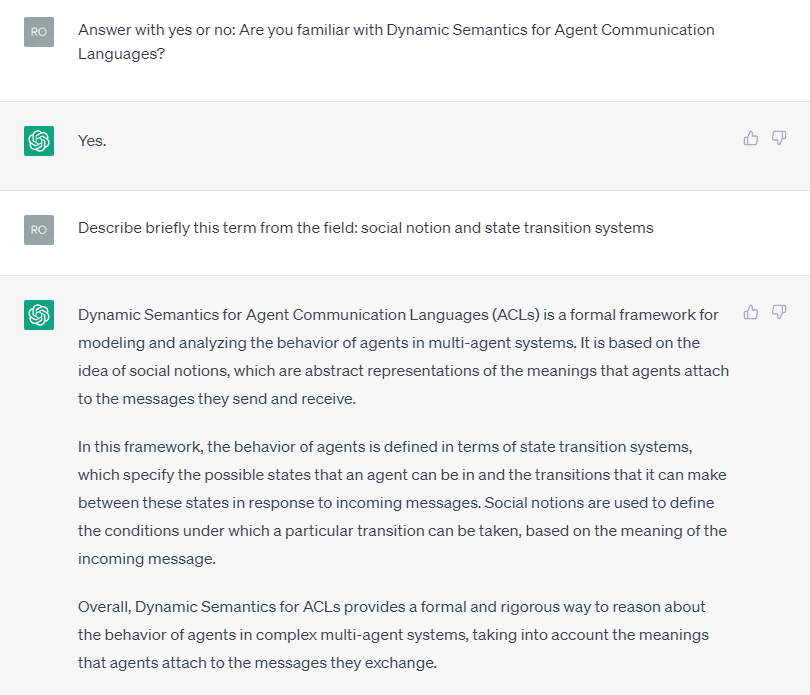}
    \caption{Case Study 2.3: Analyzing ChatGPT's Understanding of Unpredicted Keyphrases in the Document}
    \label{fig:case-study-2.3-chatgpt}
\end{figure}

In Figure \ref{fig:case-study-2.3-chatgpt}, we prompt ChatGPT to provide definitions of the keyphrases it did not predict by contextualizing the field they belong to. Surprisingly, ChatGPT can correctly recognize the field and provide accurate definitions for both phrases, despite not being able to identify them as keyphrases initially.

\subsection{Case Study 3: News Domain}
\subsubsection{Case Study 3.1}
\begin{figure}[]
    \centering
    \includegraphics[width=0.8\textwidth]{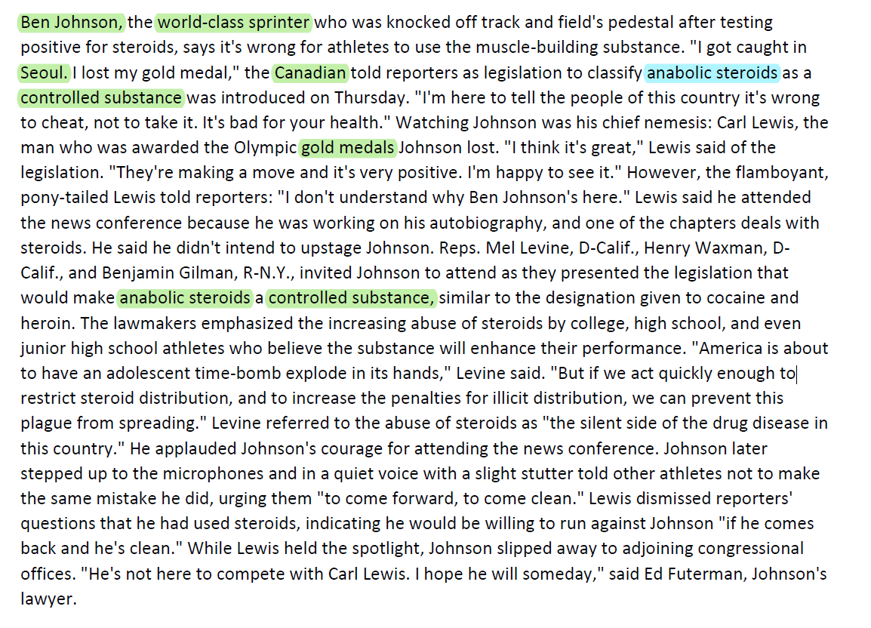}
    \caption{Case Study 3.1: News from DUC2001 Dataset}
    \label{fig:case-study-3}
\end{figure}

In this study, we assess the model's ability to generalize across domains by analyzing a sample from the DUC2001 dataset with ID 'AP891006-0029'. As depicted in Figure \ref{fig:case-study-3}, the article belongs to the sports news domain, which is markedly different from the scientific domain. This allows us to evaluate the model's capacity to handle a broad range of domains beyond scientific literature.

The drop in KeyBART's performance was expected given that the domain of the sample article is substantially different from its training data. However, the model was still able to accurately generate the scientific keyphrase 'anabolic steroids'. The model's incorrectly generated keyphrases, such as 'biomedical and behavioral research' and 'human factors', demonstrate that it still approached the article as if it were in the scientific domain, which is coherent given that this is where the model was trained. Regarding ChatGPT, its multidomain training enables it to be more robust, as it can leverage the knowledge it acquired from various tasks to generate high-quality keyphrases in any domain that was included in its training.

\begin{figure}[]
    \centering
    \includegraphics[width=0.8\textwidth]{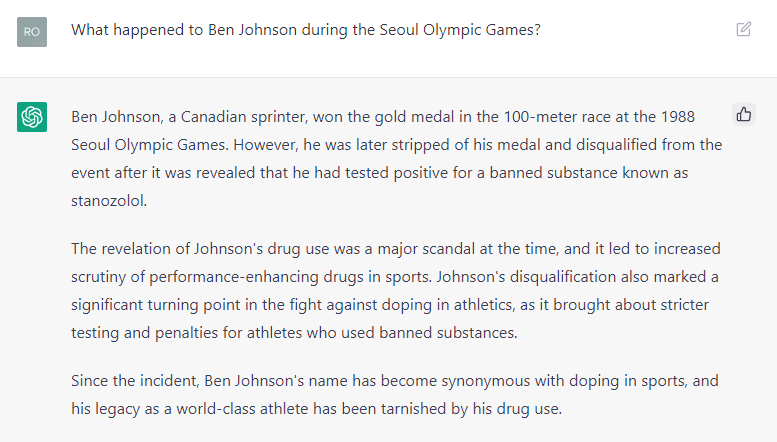}
    \caption{Case Study 3.1: ChatGPT's knowledge from the events described in the article}
    \label{fig:case-study-3-chatgpt}
\end{figure}

As the article covers an important event in sports history, we can directly test the model's knowledge of it. Figure \ref{fig:case-study-3-chatgpt} shows that ChatGPT is able to describe the events described in the article by posing a simple question, indicating that the model has latent knowledge that goes beyond the textual content. This capability enables the model to identify relevant keyphrases more accurately.

\subsubsection{Case Study 3.2}
\begin{figure}[]
    \centering
    \includegraphics[width=0.8\textwidth]{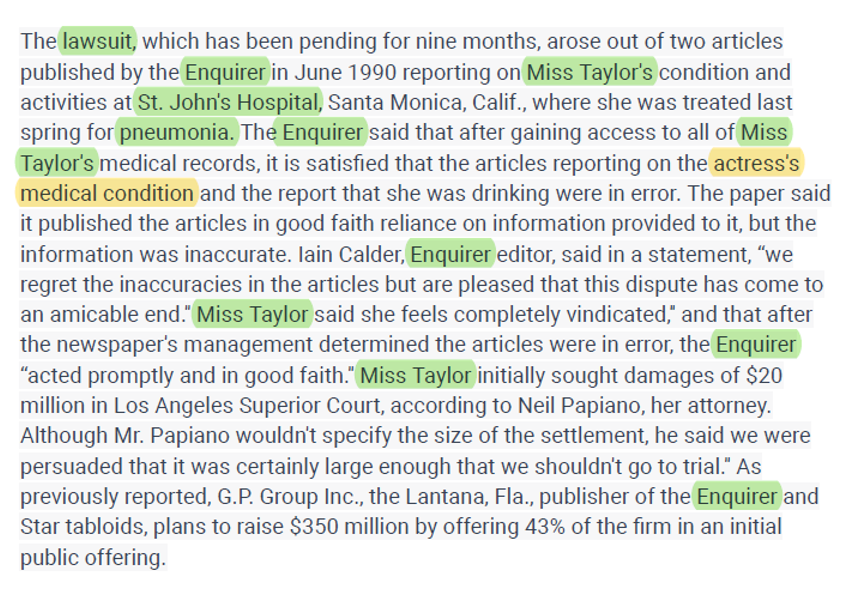}
    \caption{Case Study 3.2: News from DUC2001 Dataset}
    \label{fig:case-study-3.2}
\end{figure}

This case study evaluates the model's ability to generalize across various domains by analyzing a specific instance from the DUC2001 dataset identified as 'WSJ910529-0003'. The article, which belongs to the gossip press domain and discusses a famous actress's medical condition, is shown in Figure \ref{fig:case-study-3.2}.

As demonstrated, ChatGPT can accurately predict almost all the keyphrases, whereas KeyBART fails to predict any. KeyBART generates mostly abstractive keyphrases related to the medical domain, such as 'medical records' and 'occupational safety', which is expected since it is constrained to generate keyphrases only from the domain included in its training. Therefore, it cannot adapt well to a new domain that is far from its training data.

\begin{figure}[]
    \centering
    \includegraphics[width=0.8\textwidth]{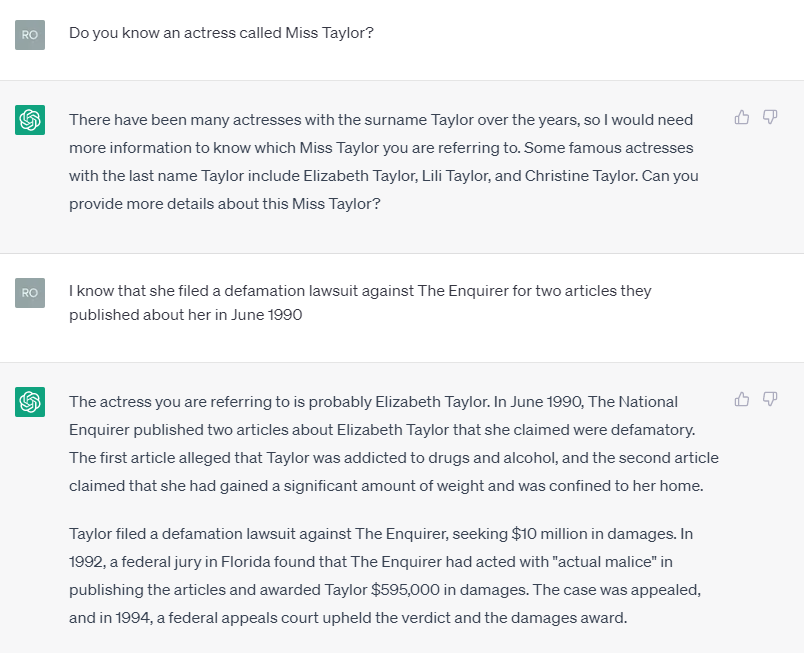}
    \caption{Case Study 3.2: ChatGPT's Knowledge of the Actress and the Events Described in the Article}
    \label{fig:case-study-3.2-chatgpt}
\end{figure}

We evaluated ChatGPT's knowledge by asking it directly about the actress and the events mentioned in the article. The results are shown in Figure \ref{fig:case-study-3.2-chatgpt}, which reveals that the model has knowledge about the events, but it struggles with details such as monetary values and locations. This knowledge is obtained from ChatGPT's pre-training on other tasks, and it significantly enhances the model's performance in the KPG task.

\subsubsection{Case Study 3.3}
\begin{figure}[]
    \centering
    \includegraphics[width=0.8\textwidth]{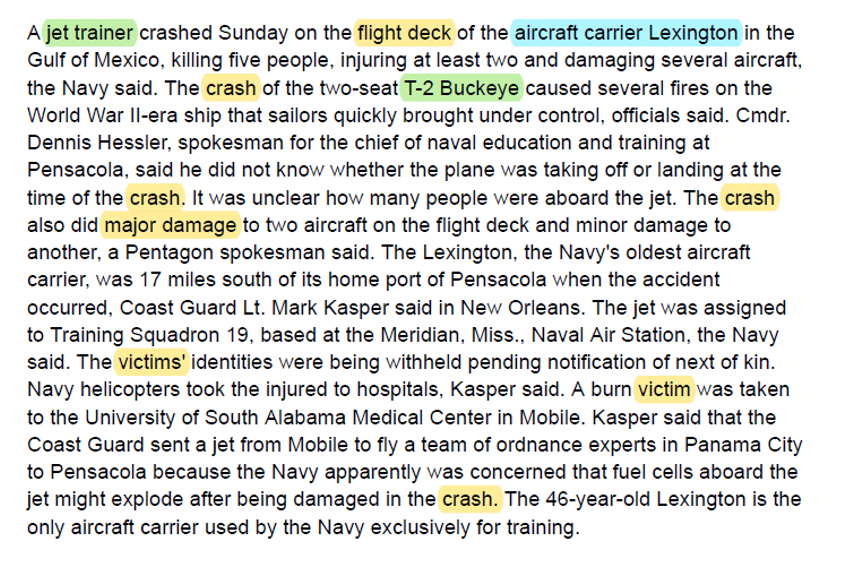}
    \caption{Case Study 3.3: News from DUC2001 Dataset}
    \label{fig:case-study-3.3}
\end{figure}

This case study examines the model's capacity to generalize across different domains by analyzing a specific instance identified as 'LA103089-0070' from the DUC2001 dataset. The article, which belongs to the aviation or military news domain, is displayed in Figure \ref{fig:case-study-3.3} and is one of the closest articles to the scientific domain that can be found in the dataset.

In this example, both models demonstrate a low level of performance, with ChatGPT's performance slightly superior to that of KeyBART's. KeyBART's predictions consist of both abstractive and extractive keyphrases from the scientific domain, indicating that the model is not resilient enough to function in an untrained domain. Some of its predictions, such as 'air traffic control' or 'suicide prevention', are not relevant to the text. The keyphrases 'crash' and 'victim' are frequently repeated, but neither of the models can identify their importance.

\begin{figure}[]
    \centering
    \includegraphics[width=0.8\textwidth]{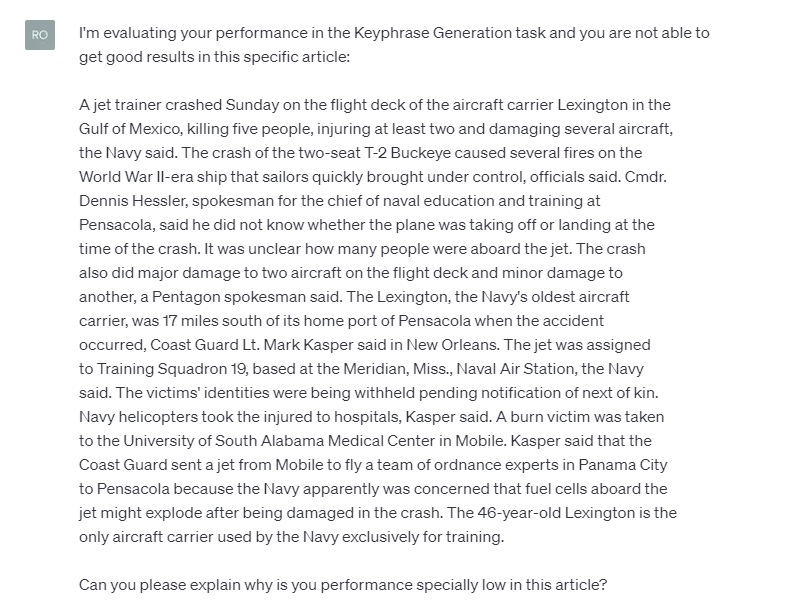}
    \caption{Case Study 3.3: Directly questioning ChatGPT about its low KPG performance in this article}
    \label{fig:case-study-3.3-chatgpt-q}
\end{figure}

\begin{figure}[]
    \centering
    \includegraphics[width=0.8\textwidth]{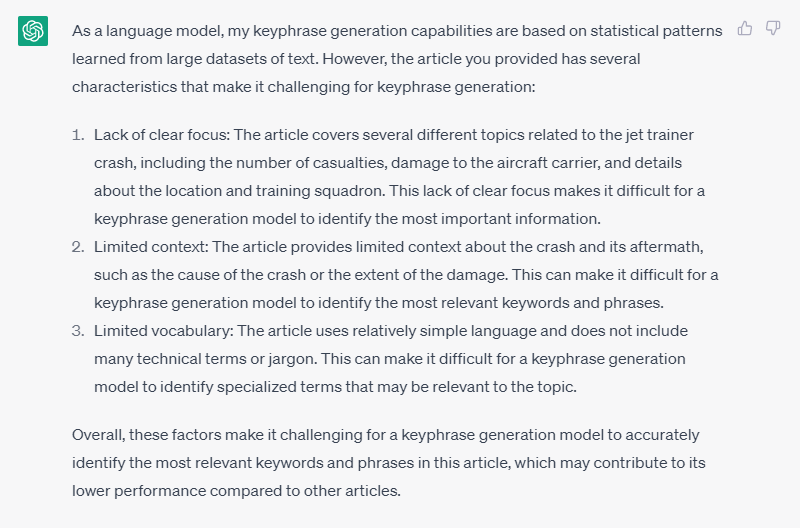}
    \caption{Case Study 3.3: ChatGPT's response to why it performed poorly in KPG in this article}
    \label{fig:case-study-3.3-chatgpt-a}
\end{figure}

In this sample, we directly asked ChatGPT to explain the reasons behind its poor performance in this specific article. The corresponding prompt and answer are depicted in Figures \ref{fig:case-study-3.3-chatgpt-q} and \ref{fig:case-study-3.3-chatgpt-a}, respectively. The model attributes its low performance to the lack of a clear focus, limited context, and limited vocabulary, which are plausible explanations that have been repeatedly identified as challenges in KPG.

%% file: conclusions.tex
As per the study, ChatGPT excels over its peers in all benchmarks, notably in handling lengthy documents and non-scientific domains. The model's superior performance is attributed to its augmented maximum input token limit and multidomain training, allowing it to gain shared in-domain knowledge from other tasks and utilize it for Keyphrase Generation. Additionally, the model benefits from a vast dataset, facilitating the training of a larger language model and increasing the maximum input tokens without any adverse impact on the KPG task's performance.

These results demonstrate that multitask and multidomain learning, with prompt-based learning playing a crucial role, can enhance keyphrase generation quality and overcome the domain adaptation and long document KPG challenges. Our study confirms that there is no specific solution for long document KPG, and all models experience significant performance degradation in this scenario. Given the increasing importance of long documents in real-world applications, it is crucial to develop more effective methods for embedding long-term relationships between words and give the model a holsitic view of the document. 